\documentclass[11pt]{article}

\usepackage[final]{acl}
\usepackage{CJKutf8}
\usepackage{times}
\usepackage{latexsym}
\usepackage{amsmath}
\usepackage{graphicx} \usepackage{booktabs}
\usepackage{multirow}
\usepackage{amssymb}
\usepackage{todonotes}
\usepackage[T1]{fontenc}

\usepackage[utf8]{inputenc}

\usepackage{microtype}

\usepackage{inconsolata}

\usepackage{graphicx}

\title{Do LLMs Capture Embodied Cognition and Cultural Variation? Cross-Linguistic Evidence from Demonstratives }

\author{Yu Wang \and Emmanuele Chersoni \and Chu-Ren Huang \\ Department of Language Science and Technology \\The Hong Kong Polytechnic University \\
        8 Hung Lok Road, Hung Hom, Kowloon, Hong Kong (China) \\
         janet-yu.wang@connect.polyu.hk,
         \{emmanuele.chersoni, churen.huang\}@polyu.edu.hk\\}         

\begin{document}
\begin{CJK}{UTF8}{gkai}
\maketitle
\begin{abstract}
Do large language models (LLMs) truly acquire embodied cognition and cultural conventions from text? We introduce demonstratives,  fundamental spatial expressions like “this/that” in English and “这/那” in Chinese, as a novel probe for grounded knowledge. Using 6,400 responses from 320 native speakers, we establish a human baseline: English speakers reliably distinguish proximal–distal referents but struggle with perspective-taking, while Chinese speakers switch perspectives fluently but tolerate distal ambiguity. In contrast, five state-of-the-art LLMs fail to inherently understand the proximal–distal contrast and show no cultural differences, defaulting to English-centric reasoning. 
Our study contributes (i) a new task, based on demonstratives, as a new lens for evaluating embodied cognition and cultural conventions; (ii) empirical evidence of cross-cultural asymmetries in human interpretation; (iii) a new perspective on the egocentric–sociocentric debate, showing both orientations coexist but vary across languages; and (iv) a call to address individual variation in future model design.
\end{abstract}

\section{Introduction}

To what extent do large language models (LLMs) acquire underlying cognitive processes and cultural conventions from their training data? LLMs have achieved remarkable progress across a wide range of linguistic tasks. While traditional narrative benchmarks have played a central role in evaluating these systems, their growing deployment in real-world applications \cite{shneiderman2020human,duan2022survey,zeng2025futuresightdrive} has created a stronger demand for testing grounded knowledge. Moreover, as LLMs operate in multilingual and multicultural contexts, it becomes essential to assess whether they can capture subtle linguistic phenomena that reflect embodied cognition and cultural variation in the languages they learn.

Demonstratives, such as “this” (proximal) and “that” (distal) in English or “这” (zhè, proximal) and “那” (nà, distal) in Chinese, provide a natural lens for such evaluation. Demonstratives encode accessibility, defined as the ease with which a speaker can obtain or interact with an object, and their interpretation varies with proximity, perspective, and cultural conventions. As a fundamental spatial expression, demonstratives are universal across languages and form part of speakers’ common ground (\citealp{himmelmann1997}; \citealp{diessel1999}; \citealp{dixon2003}; \citealp{breunesse2019}; \citealp{levinson2018}). Humans use them effortlessly to resolve referents in everyday contexts, such as distinguishing “this cup” from “that cup” when multiple similar objects are present. Children acquire the ability to use demonstratives to express accessibility between ages two and three \citep{gonzalez-pena_acquisition_2020,diessel_acquisition_2023}, highlighting their role as an early and embodied component of communication \citep{clark1978}. Moreover, accessibility is rarely made explicit in textual corpora, making it a significant challenge for LLMs trained primarily on text.

In this work, we design a bilingual dataset in English and Chinese, together with controlled experiments, to test whether LLMs can interpret demonstratives in ways consistent with human speakers. Our findings reveal that models fail to capture the proximal–distal contrast as reliably as humans, and they do not reproduce cross-linguistic differences observed in English and Chinese speakers. Specifically, English speakers show strong proximal–distal distinctions but weaker perspective-taking, while Chinese speakers emphasize perspective-taking but show weaker spatial contrast. LLMs, however, collapse these differences, reflecting the limitations of training data in capturing embodied and cultural variation.

Our work makes the following contributions:

\begin{itemize}   

\item \textbf{Novel use of demonstratives for embodied cognition.} 
We innovatively employ referent identification with demonstratives as a means of testing embodied cognition in LLMs. By focusing on demonstratives—fundamental and universal linguistic items—we move beyond surface-level text mimicry to evaluate models’ grounded knowledge.

\item \textbf{Cross-cultural evidence from human participants.} 
Based on 320 participants and 6,400 responses, we reveal cultural differences in demonstrative use, offering new empirical evidence for the ongoing debate on whether demonstratives are egocentric or socialcentric.
\item \textbf{Evaluation of LLMs across languages.} 
Our experiments show that LLMs struggle to distinguish proximal–distal contrasts and fail to reproduce human cultural differences, defaulting to English-centric reasoning despite apparent multilingual ability.

\item \textbf{Individual variation as a new challenge.} 
We highlight the challenge posed by divergent individual interpretations in referent identification. This variability exposes a limitation of current models, which are typically trained to converge on a single “expert” answer. Our findings underscore the need for individualized methods that adapt to personal differences and better support diverse users in real-world contexts.
\end{itemize}
\section{Related Work}
\subsection{Research on Demonstratives}

Studies on demonstratives have long emphasized their universality across languages, with early typological and diachronic work showing binary or ternary systems and their early emergence in child language \cite{diessel1999,himmelmann1997,dixon2003,diessel2013,todisco2020}. More recent experimental and interactional studies highlight a central debate: whether demonstratives are primarily egocentric, grounded in the speaker’s body and peripersonal space \cite{buhler1934,coventry2008,diessel2014}, or sociocentric, shaped by addressee position, shared space, and multimodal cues \cite{jarbou2010,peeters2016,rocca2019a, jara2024demonstratives}. Evidence from EEG, corpus analyses, and dialogue experiments shows that collaborative contexts can override purely spatial biases, shifting proximal space toward the partner’s action space \cite{peeters2015,rocca2019b,coventry2023}. This line of work situates demonstratives as a key site for examining the interplay between embodiment and social interaction.

\subsection{Research on LLMs and Grounded Knowledge}

LLMs have achieved strong performance in text-based tasks, yet whether they possess grounded cognition remains contested. To support richer understanding, existing research focuses on approaches such as reasoning \cite{ji2026strideedstrategygroundedstepwisereasoning, li2025mtr, wang2025think}, multimodal integration \cite{meng2026tri}, knowledge graph \cite{xue2024question}, retrieval-augmented generation \cite{zhang2026stable, lu2025deepresearch}, abstention \cite{sun2025causalabstain} etc., but they rarely probe the embodied dimensions of meaning \cite{dove2023rethinking}, grounded in bodily experience, spatial perspective, world knowledge and sensorimotor interaction \cite{wan2024perceptional}, domains that still remain challenging for LLMs because they lack physical embodiment and real‑world situatedness \citep{kauf2023event,kauf2024log,xu2025large}. 
Recent surveys stress the need for evaluation methods that go beyond accuracy to address robustness, bias, and interpretability in grounding \cite{ivanova2025,kenthapadi2024,liu2025learning, wang2024}. Against this backdrop, our study introduces demonstratives as a novel probe for grounded knowledge in LLMs, providing a simple yet powerful way to test whether models can move beyond surface-level text mimicry to capture embodied and cross-cultural aspects of meaning.

\section{Dataset Construction}
\begin{figure*}[ht]
    \centering
    \includegraphics[width=0.9\linewidth]{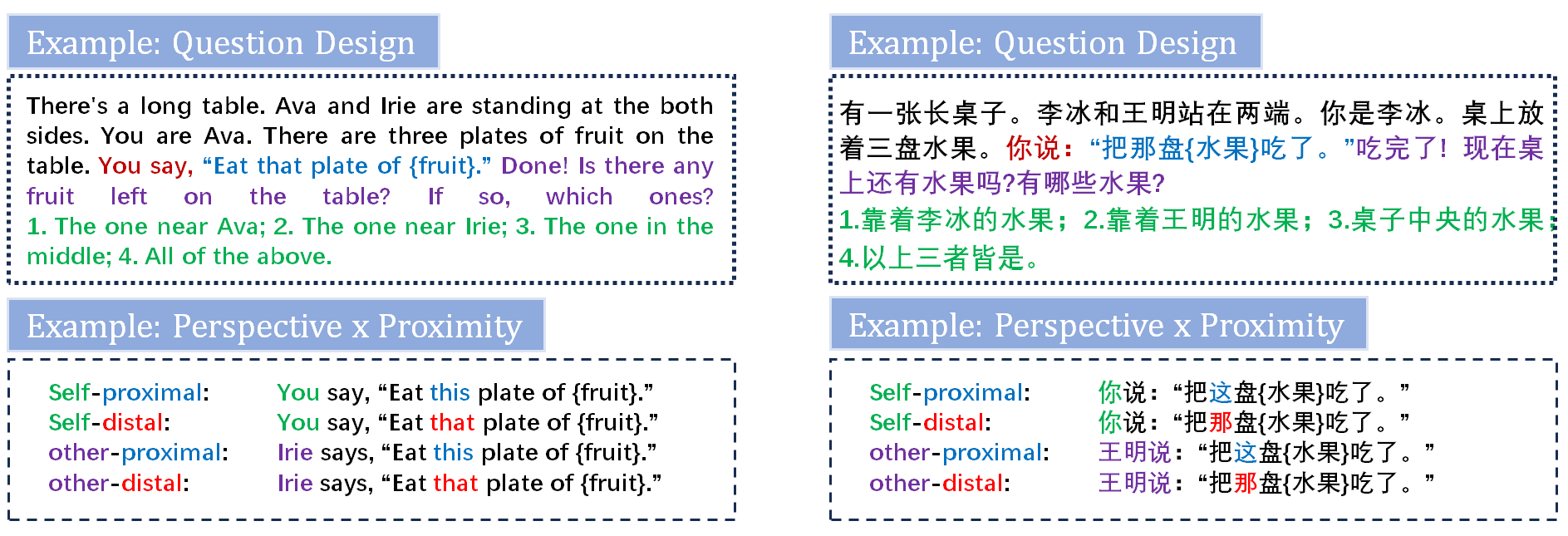}
    \caption{Example illustrating the design of data questions.}
    \label{fig:designexample}
\end{figure*}
To ensure that the evaluation goes beyond reproducing training data or guesswork, now we move to the design of the dataset.

\subsection{Question Design}
As shown in Figure \ref{fig:designexample}, each question adopts a multiple‑choice format with four options. The question begins with a description of the situation, introducing the scenario and specifying the positions of two roles who face each other from opposite sides. This is followed by the assignment of the speaker (highlighted in red) and a simple daily instruction (highlighted in blue). Within the instruction, the target object for referential disambiguation is always enclosed in curly brackets (e.g., \{fruit\} in our example).

Unlike straightforward tasks such as “which one to take”, our design poses a more complex query: “Done! Are there any {items} left on the {place}?” This requires participants or models to first identify the object referred to in the instruction, remove it conceptually, and then determine which of the remaining objects constitute the correct answer.

Finally, we design four response options to minimize guesswork. The first two represent opposites—one proximal to the speaker and one distal. The third option serves as a neutral choice, introduced to increase task complexity and reduce the likelihood of success through random guessing. The fourth option functions as a control: because selecting “All of the above” is highly improbable, repeated selection of this choice by either models or human participants can be interpreted as evidence of negating the mutual exclusivity of option 1 (proximal) and option 2 (distal). By structuring the options in this way, we broaden the response space while effectively reducing the influence of random guessing.

\subsection{Pair‑to‑Pair Settings }
Unlike traditional benchmarks, which typically assess language models using isolated questions, our dataset adopts a pair‑to‑pair design. Questions about a single scene are grouped into pairs, creating a more controlled and rigorous evaluation framework. As illustrated in Figure \ref{fig:designexample}, by systematically incorporating both proximity (proximal vs. distal) and perspective (self vs. other), each scene yields a four‑question group. This structure improves reliability and helps to ensure that performance reflects genuine reasoning rather than random guessing.

\subsection{Reference Cue Conditions}

Given that the questions involve human subjects, pronouns must be considered as an additional type of reference cue. Building on classical work on definiteness \citep{hawkins1984note, lyons1999definiteness}, we hypothesize that reinforcing pronouns strengthen referent identification by increasing definiteness and reducing ambiguity, whereas inconsistent pronouns introduce conflict and lower accuracy by disrupting the alignment between demonstrative proximity and pronoun ownership. 

Accordingly, we design four condition types: only demonstratives (this/that 这/那 zhè/nà), only pronouns (my/your 我的/你的 wǒ de/nǐ de), demonstratives+ reinforcing pronouns (this… of mine/ that... of yours 我这/你那 wǒ zhè/nǐ nà), demonstratives + inconsistent pronouns (this…of yours/ that...of mine 你这/我那 nǐ zhè/wǒ nà).  Examples for each condition are shown below:

\begin{table}[ht]
\centering
\small
\begin{tabular}{lp{0.35\linewidth}}
\hline
\textbf{Condition} & \textbf{Example} \\
\hline
Only demonstratives & \textit{Take this/that \{cup\}} \\
Only pronouns & \textit{Take my/your \{cup\}} \\
Demo + reinforcing pronouns & \textit{Take this/that \{cup\} of mine/yours} \\
Demo + inconsistent pronouns & \textit{Take that/this \{cup\} of mine/yours} \\
\hline
\end{tabular}
\caption{Four referent cue conditions.}
\end{table}

The “only demonstratives” condition serves as the primary experimental baseline, reflecting the most common way objects are referenced in everyday communication. The other three conditions act as controls, varying pronoun use to examine its impact on referent identification.

\subsection{Statistics of the Dataset}
Building on the above design, we develop our dataset within a controlled experimental framework that integrates two proximity conditions, two perspectives, and four types of referent cues. To further ensure robustness, we introduce five everyday scenarios with varied verbs (e.g., \textit{eat, hide, take}), enabling us to examine if and to what extent situational context and verb semantics affect performance. Notably, unlike typical NLP studies that emphasize scale, our focus is on carefully designed instances. The dataset comprises 80 questions per language, totaling 160.

\begin{figure*}[ht]
    \centering
    \includegraphics[width=0.83\linewidth]{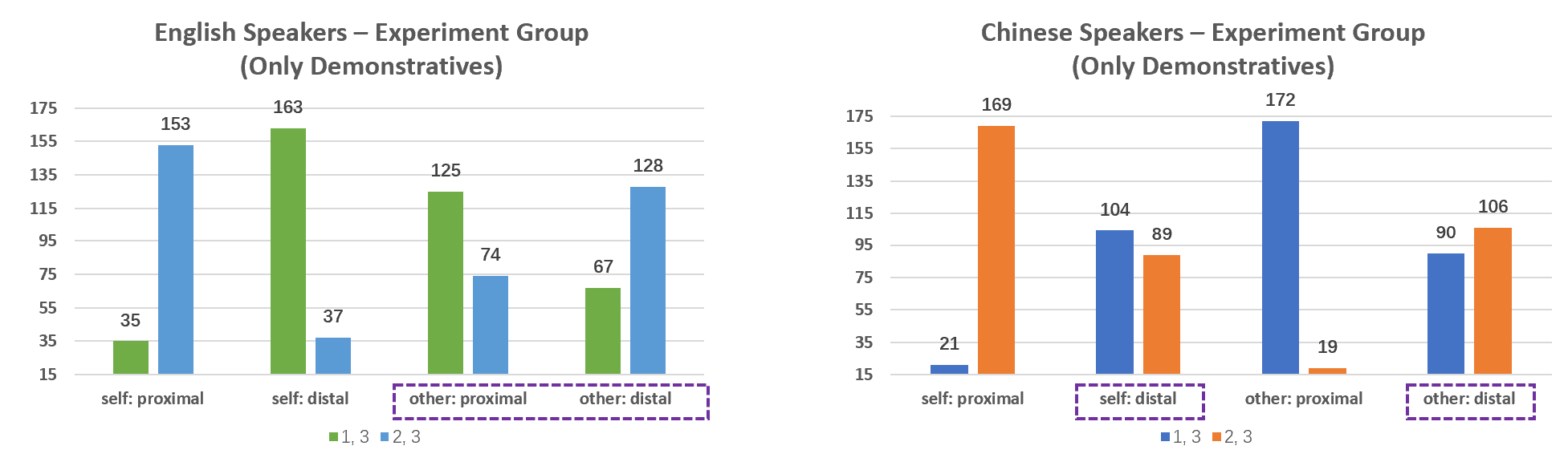}
    \caption{Distribution of answers across conditions in the experimental group. Purple dashed boxes highlight categories  with divergent responses. Further details on alternative choices are provided in the Appendix.}
    \label{fig:experiment_result}
\end{figure*}

\begin{table*}[ht]
\centering
\small
\begin{tabular}{llcc}
\hline
 \textbf{Comparison} &  &\textbf{English SI} & \textbf{Chinese SI} \\
\hline
 Self Proximal &  vs.  Self Distal & \textbf{0.0309 $\checkmark$} & 0.3472 $\times$ \\
 Other Proximal &  vs.  Other Distal & \textbf{0.0254 $\checkmark$} & 0.3541 $\times$ \\
 Self Proximal &  vs.  Other Proximal & 0.1731 $\times$ & \textbf{0.0131 $\checkmark$} \\
 Self Distal &  vs.  Other Distal & 0.1646 $\times$ & \textbf{0.0077 $\checkmark$} \\
\hline
\end{tabular}
\caption{Symmetry Index (SI) results for English and Chinese speakers. Threshold = 0.1: values below (bold) indicate high symmetry, above indicate asymmetry. English speakers show reliable proximal–distal distinctions but weaker self–other switching, while Chinese speakers excel in self–other switching yet are less consistent in proximal–distal distinctions.}
\label{tab:SI_results}
\end{table*}

\section{Human Performance}

To avoid cross‑condition influence between different reference cues, for example, participants exposed to only demonstratives might be biased when later answering questions with demonstratives combined with pronouns, we divide the dataset into four separate surveys, each restricted to a single reference‑cue type and comprising 20 questions. The questions are randomly shuffled to minimize order effects. Each survey include 40 participants (balanced by gender), resulting in a total of 320 native speakers: 160 Chinese and 160 English. The Chinese surveys are conducted on Credamo, and the English surveys on Prolific. In total, the study collect \textbf{6,400} question responses.

\subsection{Results of Experiment Group}
The experimental group consists of conditions in which demonstratives (e.g., this/that) are used as the sole referential cues. Referent identification is known to rely heavily on contextual information. In the absence of sufficient context, multiple interpretations might arise, resulting in divergent answers to the same item. Consequently, we do not designate a single “correct” answer for evaluation. Instead, we analyze human performance by examining the distribution of responses.

\subsubsection{Performance on Proximal–Distal Contrasts}
As shown in Figure \ref{fig:experiment_result}, the results confirm prior findings that speakers across languages share a robust understanding of the proximal–distal distinction. Both Chinese and English participants consistently recognize this contrast. Importantly, they rarely select directly opposing options—one proximal and one distal (options 1 and 2). Instead, responses reveal a systematic tendency to combine a proximal choice with the middle option (1, 3), or a distal choice with the middle option (2, 3).  These dominant response patterns account for 97.13\% of all answers, underscoring the stability and universality of the proximal–distal contrast in human interpretation. 

\subsubsection{Performance on Perspective-Taking and Proximity}

An examination of perspective-taking and proximity reveals striking cross-linguistic differences. To quantify these patterns, we decide to utilize the Symmetry Index (SI) \cite{Robinson1987SI}, a normalized measure of how balanced two paired conditions are:

\begin{equation}
\footnotesize
\text{SI} = 
\frac{\left| A_1 - B_2 \right| 
+ \left| B_1 - A_2 \right|}
{A_1 + A_2 + B_1 + B_2}
\end{equation}

where $A_1, A_2$ and $B_1, B_2$ denote the counts of two response categories under comparison. Lower SI values indicate stronger symmetry.

\textbf{English Speakers.}  
English speakers exhibit strong symmetry within a single perspective: both Self Proximal vs. Self Distal (SI = 0.0309) and Other Proximal vs. Other Distal (SI = 0.0254) fall well below the 0.1 threshold, indicating consistent proximal–distal distinctions. However, symmetry breaks down across perspectives: Self Proximal vs. Other Proximal (SI = 0.1731) and Self Distal vs. Other Distal (SI = 0.1646) exceed the threshold, revealing substantial asymmetry. This divergence is also evident in Figure \ref{fig:experiment_result}: when the perspective shift to “other”, responses become more variable. Roughly two-thirds of participants adopt the interlocutor’s viewpoint, while one-third continue to respond from their own. These results indicate that English speakers excel at distinguishing proximal vs. distal demonstratives within a self perspective, but encounter significant difficulty when shifting to other viewpoints.

\textbf{Chinese Speakers.}  
Chinese speakers show the opposite pattern. They demonstrate strong symmetry across perspectives: Self Proximal vs. Other Proximal (SI = 0.0131) and Self Distal vs. Other Distal (SI = 0.0077) indicate fluent other perspective-taking. However, facing "distal" issues, symmetry collapses. Self Proximal vs. Self Distal (SI = 0.3472) and Other Proximal vs. Other Distal (SI = 0.3541) suggest inconsistent interpretation of demonstrative contrasts. As shown in Figure \ref{fig:experiment_result}, responses to distal forms such as 那 (\textit{nà}, “that”) are evenly split—about half interpret the referent as near the interlocutor, while the other half treat it as near themselves. This ambiguity complicates referent identification and undermines proximal–distal consistency.

Taken together, the results reveal a complementary cross-linguistic pattern:  

\textbf{English:} strong proximal–distal distinction, but weak perspective-taking beyond the self.   

\textbf{Chinese:} strong perspective-taking beyond the self, but weak proximal–distal distinction.  

This contrast underscores that demonstrative interpretation is shaped not only by spatial encoding, but also by cultural-linguistic conventions.

\subsection{Results of Control Group}
\begin{figure*}[ht]
    \centering
    \includegraphics[width=1\linewidth]{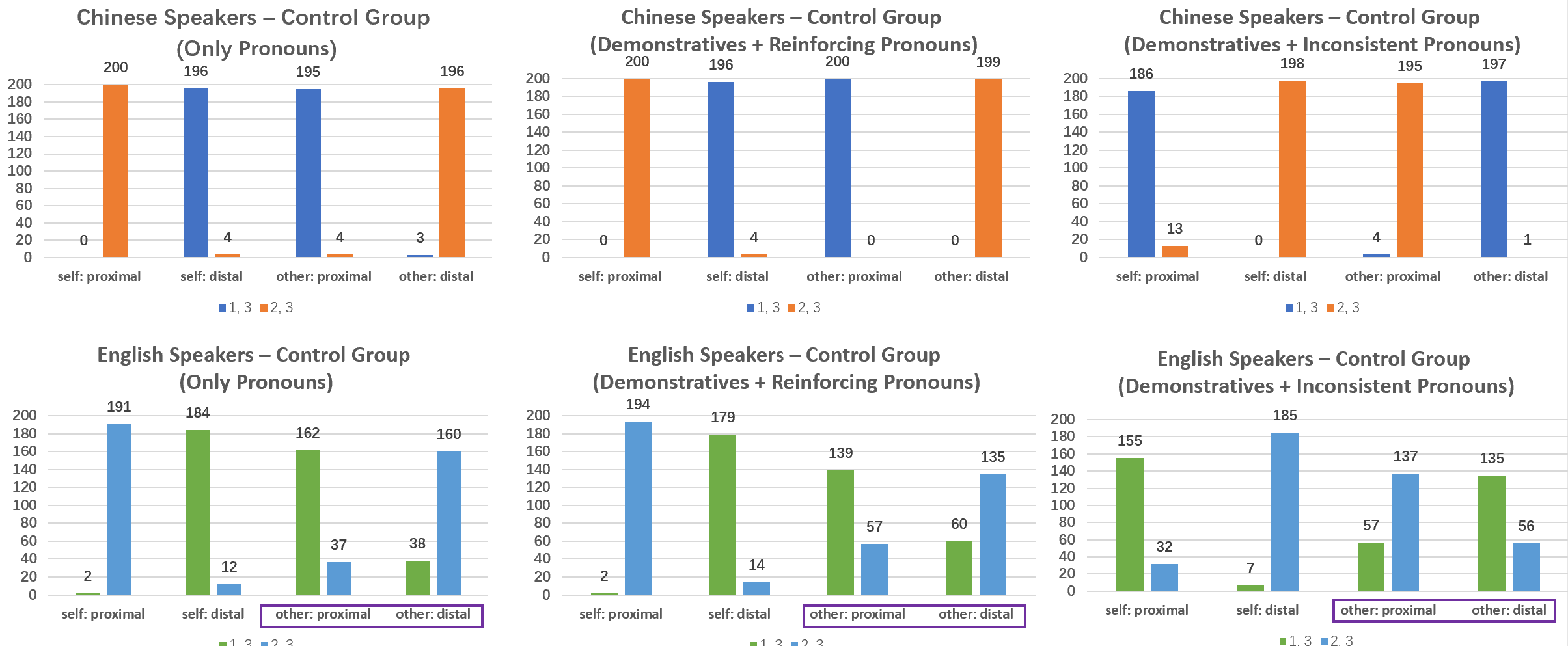}
    \caption{Distribution of answers across conditions in the control group}
    \label{fig:control_data}
\end{figure*}

The control group comprises three conditions: only pronouns, demonstratives with reinforcing pronouns, and demonstratives with inconsistent pronouns. 

In the only pronoun condition (Figure \ref{fig:control_data}, left), participants rely primarily on pronouns, yielding highly consistent responses. When demonstratives are paired with reinforcing pronouns (Figure \ref{fig:control_data}, middle), referent identification becomes clearer and more stable compared with the only demonstrative condition (Figure \ref{fig:experiment_result}). Reinforcing pronouns increase definiteness and reduce ambiguity, strengthening the alignment between demonstrative proximity and pronoun ownership. By contrast, inconsistent pairings (Figure \ref{fig:control_data}, right) produce interpretive uncertainty. Participants consistently privilege pronouns over demonstratives, even when the two conflict. The distributions in reinforcing and inconsistent conditions are nearly opposite, underscoring the overriding influence of pronouns.

Overall, the results confirm our hypothesis: reinforcing pronouns facilitate referent identification, while inconsistent pronouns disrupt alignment and reduce accuracy. 

Across conditions, pronouns consistently dominate referent identification. With pronouns, Chinese speakers overcome distal ambiguity and remain stable across proximities and perspectives. English speakers are reliable in self-perspective but variable in interlocutor-perspective. These findings parallel the experimental group: Chinese speakers switch perspectives fluently but face conflicts that arise from distal ambiguity, whereas English speakers struggle with perspective-taking beyond the self.

\subsection{Discussion}

The observed cross-linguistic differences raise the question of why Chinese and English speakers exhibit such differences in demonstrative interpretation. One plausible explanation lies in the interaction between linguistic structure and cultural conventions. 

It is widely acknowledged in cross-cultural communication research that English and Chinese exhibit different communicative styles. One influential framework is Hall’s theory of contexting, according to which English is typically classified as a low-context language, favoring explicitness and precision, whereas Chinese is a high-context language, allowing for ambiguity and relying on shared context (\citealp{Hall1976}; \citealp{Gudykunst1988}; \citealp{Samovar2004}). A familiar illustration is recipe language: English specifies “200 grams of sugar”, while Chinese often uses “适量” (shì liàng, an appropriate amount) leaving interpretation room. This distinction parallels demonstrative usage, where English speakers maintain precise proximal–distal contrasts while Chinese speakers rely more on contextual flexibility, producing variability in distal interpretation.

With respect to perspective switching, as discussed previously, considerable debate exists over whether demonstrative systems are egocentric—anchored in the speaker’s body and viewpoint—or sociocentric—co-constructed through alignment with the interlocutor. Our findings suggest that both orientations are present, but the dominant tendency varies across languages. For English speakers, demonstrative interpretation shows a strong egocentric orientation: the system places precise emphasis on proximal–distal contrasts, yielding high consistency within a single perspective. However, when required to adopt the interlocutor’s viewpoint, reliance on self-centered encoding produces asymmetry and variability. By contrast, Chinese speakers exhibit a more sociocentric orientation, often demonstrating sensitivity to the interlocutor’s perspective, which explains the strong symmetry observed across perspectives.

The control group results further support this account. When pronouns were introduced, Chinese speakers overcome distal ambiguity and maintain stability across perspectives, while English speakers remain reliable in self-perspective but variable in interlocutor-perspective. Taken together, the evidence highlights that demonstrative interpretation is not governed solely by spatial reasoning. Instead, it reflects deeper cultural-linguistic orientations: English favors clear proximity distinctions but struggles with perspective shifts, whereas Chinese favors flexible perspective-taking but allows for ambiguity in proximity contrasts. This complementary pattern raises a critical question: Do large language models perform differently depending on the language they use? Beyond performance, do they also acquire the specific communicative styles (i.e. egocentric vs. sociocentric orientation) embedded in the training data of that language?
\begin{table*}[htbp]
\centering

\resizebox{\textwidth}{!}{%
\begin{tabular}{lcccccccccccccccc}
\toprule
\textbf{Model} & \multicolumn{8}{c}{\textbf{Gpt5.1}} & \multicolumn{8}{c}{\textbf{Claude-sonnet-4.5}} \\
\cmidrule(lr){2-9}\cmidrule(lr){10-17}
\textbf{Class} &
\multicolumn{2}{c}{\textbf{Self: Proximal}} & \multicolumn{2}{c}{\textbf{Self: Distal}} & \multicolumn{2}{c}{\textbf{Other: Proximal}} & \multicolumn{2}{c}{\textbf{Other: Distal}} &
\multicolumn{2}{c}{\textbf{Self: Proximal}} & \multicolumn{2}{c}{\textbf{Self: Distal}} & \multicolumn{2}{c}{\textbf{Other: Proximal}} & \multicolumn{2}{c}{\textbf{Other: Distal}} \\
\textbf{Lang} & \textbf{en} & \textbf{zh} & \textbf{en} & \textbf{zh} & \textbf{en} & \textbf{zh} & \textbf{en} & \textbf{zh} &
\textbf{en} & \textbf{zh} & \textbf{en} & \textbf{zh} & \textbf{en} & \textbf{zh} & \textbf{en} & \textbf{zh} \\
\midrule
\textbf{1,3} & / & / & 30.00\% & 24.00\% & 60.00\% & 70.00\% & / & / & / & 2.00\% & 34.00\% & 6.00\% & 48.00\% & / & 4.00\% & / \\
\textbf{2,3} & 72.00\% & 80.00\% & / & / & / & / & 30.00\% & 48.00\% & 48.00\% & / & / & / & 16.00\% & / & 24.00\% & / \\
\textbf{4} & 20.00\% & 18.00\% & 40.00\% & 46.00\% & 26.00\% & 26.00\% & 46.00\% & 32.00\% & 48.00\% & 30.00\% & 44.00\% & 66.00\% & 12.00\% & 40.00\% & 52.00\% & 76.00\% \\
\textbf{1} & / & 2.00\% & 24.00\% & 14.00\% & 10.00\% & 2.00\% & / & 4.00\% & / & 14.00\% & 4.00\% & 14.00\% & 14.00\% & / & / & 6.00\% \\
\textbf{2} & 8.00\% & / & / & 2.00\% & / & / & 6.00\% & 4.00\% & / & 54.00\% & 6.00\% & 14.00\% & / & 60.00\% & 12.00\% & 12.00\% \\
\textbf{3} & / & / & / & / & / & / & / & / & / & / & / & / & 2.00\% & / & 2.00\% & / \\
\textbf{1,2} & / & / & / & 14.00\% & / & / & 2.00\% & 8.00\% & 2.00\% & / & 10.00\% & / & 6.00\% & / & 2.00\% & 6.00\% \\
\textbf{None} & / & / & 6.00\% & / & 4.00\% & 2.00\% & 16.00\% & 4.00\% & 2.00\% & / & 2.00\% & / & 2.00\% & / & 4.00\% & / \\
\bottomrule
\end{tabular}
}

\vspace{0.5em}

\resizebox{\textwidth}{!}{%
\begin{tabular}{lcccccccccccccccc}
\toprule
\textbf{Model} & \multicolumn{8}{c}{\textbf{Gemini-2.5-pro}} & \multicolumn{8}{c}{\textbf{Deepseek-v3.1}} \\
\cmidrule(lr){2-9}\cmidrule(lr){10-17}
\textbf{Class} &
\multicolumn{2}{c}{\textbf{Self: Proximal}} & \multicolumn{2}{c}{\textbf{Self: Distal}} & \multicolumn{2}{c}{\textbf{Other: Proximal}} & \multicolumn{2}{c}{\textbf{Other: Distal}} &
\multicolumn{2}{c}{\textbf{Self: Proximal}} & \multicolumn{2}{c}{\textbf{Self: Distal}} & \multicolumn{2}{c}{\textbf{Other: Proximal}} & \multicolumn{2}{c}{\textbf{Other: Distal}} \\
\textbf{Lang} & \textbf{en} & \textbf{zh} & \textbf{en} & \textbf{zh} & \textbf{en} & \textbf{zh} & \textbf{en} & \textbf{zh} &
\textbf{en} & \textbf{zh} & \textbf{en} & \textbf{zh} & \textbf{en} & \textbf{zh} & \textbf{en} & \textbf{zh} \\
\midrule
\textbf{1,3} & / & / & 10.00\% & 10.00\% & 36.00\% & 36.00\% & / & / & / & / & / & / & 10.00\% & 2.00\% & 6.00\% & / \\
\textbf{2,3} & 62.00\% & 48.00\% & / & / & / & / & 16.00\% & 10.00\% & 22.00\% & 4.00\% & 6.00\% & 2.00\% & 2.00\% & 2.00\% & 4.00\% & 2.00\% \\
\textbf{4} & 36.00\% & 18.00\% & 42.00\% & 28.00\% & 58.00\% & 48.00\% & 56.00\% & 46.00\% & 44.00\% & 48.00\% & 56.00\% & 60.00\% & 40.00\% & 68.00\% & 44.00\% & 68.00\% \\
\textbf{1} & / & / & 48.00\% & 52.00\% & 4.00\% & 4.00\% & / & / & 2.00\% & 12.00\% & / & 10.00\% & / & 8.00\% & / & 14.00\% \\
\textbf{2} & 2.00\% & 20.00\% & / & 2.00\% & / & 2.00\% & 22.00\% & 34.00\% 
           & 16.00\% & 14.00\% & / & 14.00\% & 2.00\% & 8.00\% & 4.00\% & 10.00\% \\
\textbf{3} & / & 2.00\% & / & 2.00\% & / & 2.00\% & 2.00\% & / 
           & 2.00\% & 22.00\% & / & 4.00\% & / & 8.00\% & / & 4.00\% \\
\textbf{1,2} & / & 2.00\% & / & 6.00\% & / & / & 2.00\% & 4.00\% 
             & 12.00\% & / & 38.00\% & 10.00\% & 44.00\% & 2.00\% & 42.00\% & 2.00\% \\
\textbf{None} & / & 10.00\% & / & / & 2.00\% & 8.00\% & 2.00\% & 6.00\% 
              & 2.00\% & / & / & / & 2.00\% & 2.00\% & / & / \\
\bottomrule
\end{tabular}
}

\vspace{0.5em}

\resizebox{\textwidth}{!}{%
\begin{tabular}{lcccccccccccccccc}
\toprule
\textbf{Model} & \multicolumn{8}{c}{\textbf{Qwen3-max}} & \multicolumn{8}{c}{\textbf{Human}} \\
\cmidrule(lr){2-9}\cmidrule(lr){10-17}
\textbf{Class} &
\multicolumn{2}{c}{\textbf{Self: Proximal}} & \multicolumn{2}{c}{\textbf{Self: Distal}} & \multicolumn{2}{c}{\textbf{Other: Proximal}} & \multicolumn{2}{c}{\textbf{Other: Distal}} &
\multicolumn{2}{c}{\textbf{Self: Proximal}} & \multicolumn{2}{c}{\textbf{Self: Distal}} & \multicolumn{2}{c}{\textbf{Other: Proximal}} & \multicolumn{2}{c}{\textbf{Other: Distal}} \\
\textbf{Lang} & \textbf{en} & \textbf{zh} & \textbf{en} & \textbf{zh} & \textbf{en} & \textbf{zh} & \textbf{en} & \textbf{zh} &
\textbf{en} & \textbf{zh} & \textbf{en} & \textbf{zh} & \textbf{en} & \textbf{zh} & \textbf{en} & \textbf{zh} \\
\midrule
\textbf{1,3} & / & / & / & / & / & / & / & / & 17.50\% & 10.50\% & 81.50\% & 52.00\% & 62.50\% & 86.00\% & 33.50\% & 45.00\% \\
\textbf{2,3} & 20.00\% & 12.00\% & 12.00\% & 2.00\% & 2.00\% & 4.00\% & 2.00\% & 2.00\% & 76.50\% & 84.50\% & 18.50\% & 44.50\% & 37.00\% & 9.50\% & 64.00\% & 53.00\% \\
\textbf{4} & 80.00\% & 54.00\% & 84.00\% & 74.00\% & 76.00\% & 70.00\% & 80.00\% & 64.00\% & 1.00\% & / & / & / & / & / & / & / \\
\textbf{1} & / & / & / & / & / & / & / & / & 0.50\% & 0.50\% & / & 0.50\% & / & 0.00\% & 0.50\% & 0.00\% \\
\textbf{2} & / & 12.00\% & / & 6.00\% & / & / & / & 2.00\% & / & 0.50\% & / & 0.00\% & / & 0.50\% & / & 0.50\% \\
\textbf{3} & 16.00\% & 4.00\% & 14.00\% & 18.00\% & 18.00\% & 18.00\% & 26.00\% & / & 1.50\% & 3.00\% & / & 1.50\% & / & 3.00\% & 1.50\% & 1.00\% \\
\textbf{1,2} & / & 6.00\% & / & 4.00\% & 4.00\% & 8.00\% & / & 6.00\% & 3.00\% & 1.00\% & / & 1.50\% & / & 0.50\% & 0.50\% & 0.50\% \\
\textbf{None} & / & / & / & / & / & / & / & / & / & / & / & / & / & / & / & / \\
\bottomrule
\end{tabular}
}
\caption{Comparison across models and human. The results reported are averages from 10 runs. \textit{en} denotes English, while \textit{zh} denotes Chinese. }
\label{tab:experiment_result}
\end{table*}

\section{Model Evaluation}

Our evaluation includes three closed‑source models—GPT‑5.1 (2025.11.12) \cite{openai2025gpt51}, Claude 4.5 Sonnet (2025.09.29) \cite{anthropic2025claude45}, and Gemini 2.5 Pro (2025.06.17) \cite{deepmind2025gemini25pro}—as well as two recent open‑source models: Deepseek V3.1 (2025.08.21) \cite{deepseek2025v31} and Qwen3 Max (2025.09.29) \cite{yang2025qwen3} \footnote{Model selection is based on the versions publicly available as of 2025‑11‑15.}. Drawing on the comparative analysis (see Appendix for details), we adopt a zero-shot prompt refinement strategy for all subsequent experiments. Each model receives only the question text and a standardized instruction: “Please reply [The answer is: {1, 2, 3, 4}], and give a brief reason.” No additional context or exemplars are supplied, ensuring methodological consistency and minimizing external bias. To obtain stable estimates, each model is run ten times under identical conditions, yielding a total of 4,800 test instances. We compare each run with the overall distribution and observe only small fluctuations, with standard deviations typically between 0.02 and 0.08 and only occasional outliers reaching 0.12, indicating that the models produce relatively consistent answers across runs. 

To statistically assess whether model responses deviate from human behavior, we apply the Rao–Scott adjusted chi‑square test to the answer distributions of the experiment group questions between humans and models (this test is a classical choice for dealing with multiple response categorical variables, as in the case of our data). Across all models and both languages, the vast majority of items show significant divergence from human distributions, with most comparisons yielding p < .05, except for the comparison human vs. models on the self-perspective proximal condition, indicating strong statistical confidence that model outputs do not originate from the same underlying generative process as human judgments. Jensen–Shannon divergence provides a complementary measure of effect size, with mean values ranging from 0.30 to 0.66, reflecting substantial distributional distance. The only conditions where models approach human‑like behavior are the self‑perspective proximal items, where several systems produce more human‑aligned distributions and occasionally yield higher p‑values and lower JS divergence. 

\subsection{Evaluation Result of Experiment Group}

Based on the complexity of human responses, we evaluate models not by accuracy but by comparing the distribution of their responses with those of humans. The left column in Table \ref{tab:experiment_result} presents the choices made by the models among the four options (e.g., 1,3 / 2,3). As the table shows, all models perform poorly on referent identification with demonstratives, producing results that differ significantly from human patterns and remain inconsistent even among themselves. 

\textbf{Do LLMs understand the contrast between proximal and distal demonstratives as human participants do?}

The answer is no. With an intuitive understanding that proximal (option 1) and distal (option 2) demonstratives are mutually exclusive categories, human participants demonstrate a clear understanding of the task: almost no one (approximately 0.5\% on average) has selected option 4 (“All of the above”) or the logically incoherent combination of option 1 and 2. 

By contrast, models frequently default to option 4. For instance, Gemini‑2.5‑pro selects option 4 in 36–56\% of English cases and up to 60\% in Chinese, while Deepseek‑v3.1 shows even higher reliance, reaching 44–68\% across conditions. One might argue that this reflects a conservative answering strategy rather than misunderstanding. However, that is precisely the point: when models fail to recognize that the task specifically targets demonstrative contrasts (which is the very purpose of our dataset design), they no longer primarily mimic human performance on demonstratives. Instead, their behavior shifts toward imitating human answering patterns in other domains, such as generic puzzle strategies. In adopting this “safe fallback” to maximize performance, they often endorse options that are logically incompatible. Such behavior reveals that models do not genuinely treat proximal and distal references as mutually exclusive. Unlike humans, who avoid combining incompatible options even under uncertainty, models’ reliance on option 4 underscores their lack of genuine understanding of the proximal–distal contrast.

\textbf{Do LLMs perform culturally differently across Chinese and English?}

The answer is no. As discussed in the human performance section and shown in Table \ref{tab:experiment_result}, Chinese participants encounter ambiguity when interpreting distal references, leading to a split between option 2,3 (52.0\%) and option 1,3 (44.5\%) in the self–distal condition. In contrast, English speakers exhibit greater ease in interpreting distal demonstratives, showing a strong preference for option 2,3 (81.5\%) and option 1,3 (18.5\%) .

At the same time, Chinese speakers demonstrate fluent perspective switching between self and other conditions. For instance, in the other–proximal condition, 86.0\% select option 1,3 while only 9.5\% choose option 2,3. English participants, however, struggle with perspective switching, displaying less consistency: 62.5\% select option 1,3, whereas 37.0\% opt for 2,3—again reflecting weaker perspective-shifting ability.

In contrast, LLMs do not exhibit culturally distinct patterns. Their response distributions across Chinese and English remain largely uniform, suggesting that they fail to internalize language-specific pragmatic tendencies. For example, Gemini‑2.5‑pro shows highly similar results across the two languages: in four conditions, its choices for (2,3 / 2 or 1,3 / 1) are 64\%, 58\%, 40\%, and 38\% in English, compared with 68\%, 62\%, 40\%, and 44\% in Chinese. The differences range only from 0\% to 6\%, indicating no substantial divergence between the two languages. Although these models are described as multilingual, they do not demonstrate language-specific modes of reasoning. A closer examination reveals that they perform better in self-perspective conditions but worse in other-perspective conditions across both languages—patterns that align more closely with English usage. It is possible that, even when responding in Chinese, the models tend to adopt an English-oriented way of reasoning rather than fully reflecting culturally specific interpretive patterns.

\subsection{Evaluation Result of Control Group}
The results of the control group further support the findings from the human performance and experimental group (due to space limitations, full data are provided in the Appendix). Consistent with our hypothesis, models perform better when pronouns are used, as they provide clearer referential cues. For instance, in demonstratives with reinforcing pronouns, Claude-sonnet-4.5 selects option 1,3 in 80.0\% of English self–distal pronoun cases, closely aligning with human performance (89.5\% option 1,3), compared with only 34.0\% in only demonstrative conditions. When facing inconsistent pronouns, models also tend to follow pronouns rather than demonstratives. Although the models perform better overall, they still show a strong tendency to select option 4 or option 1,2—two choices that should be mutually exclusive—highlighting their lack of genuine understanding of the proximal–distal contrast. 

Regarding cultural variation, models do not exhibit language-specific differences. For example, in reinforcing pronoun + demonstrative conditions, English speakers continue to struggle with other-perspective interpretation (e.g., in the self–distal condition, 89.5\% select 1/1,3, whereas in the other–distal condition only 67.5\% select 2/2,3). By contrast, models such as Claude-sonnet-4.5 show nearly identical distributions across languages (e.g., self–distal: 94.0\% select 1/1,3 and other–distal: 94.0\% select 2/2,3). This suggests that LLMs, despite their multilingual nature, do not internalize language-specific pragmatic patterns.

\section{Conclusion}

In this paper, we introduce demonstratives as a novel probe for evaluating embodied cognition and cultural variation in large language models. Using a carefully constructed dataset that systematically varies proximity, perspective, and reference cues, and drawing on 6,400 responses from 320 native speakers, we establish a robust human baseline. This design reveals complementary cultural patterns: English speakers emphasize proximal–distal contrasts but struggle with perspective-taking, whereas Chinese speakers switch perspectives fluently but tolerate distal ambiguity. Our findings contribute to the ongoing debate over whether demonstrative systems are egocentric or sociocentric, showing that both orientations are present but that their dominance varies across languages.

Against this baseline, our evaluation of five state‑of‑the‑art LLMs demonstrates that models do not genuinely understand demonstratives. Instead, they mimic surface‑level answering strategies, often defaulting to logically incompatible options such as “All of the above”. Pronouns provide clearer cues and improve performance, yet models still fail to internalize the mutual exclusivity of proximal and distal references. Crucially, they show no cultural differences: distributions remain uniform across English and Chinese, reflecting an English‑centric reasoning style rather than culturally grounded interpretation.

Finally, our study highlights individual variation as a new challenge. Human interpretations of demonstratives are diverse, but current models are trained to converge on a single “expert” answer. Addressing these gaps will require systems that move beyond mimicry to incorporate embodied, culturally specific, and individualized reasoning. Demonstratives thus offer a powerful lens for probing grounded cognition, exposing the limits of current LLMs, and pointing toward the need for models that can better support diverse users in real‑world contexts.

\section*{Limitations}
Despite the contributions, our study has several limitations that point to directions for future work.

\textbf{Embodied grounding:} Our current study is limited to text-only input, which cannot fully capture the embodied nature of demonstrative use. Referential interpretation in real communication relies on multi-dimensional inputs, spatial orientation, visual perception, and physical interaction, rather than abstract text alone. To approximate the embodied reality of demonstratives, future work should incorporate multimodal datasets (combining text, images, and possibly audio) and explore 3D simulation environments where agents can interact with objects and perspectives in space \cite{wang2026one}. Advancing in this direction will be crucial for developing embodied AI and robotic systems that can engage in natural referential reasoning and adapt to diverse human communicative contexts.

\textbf{Dataset scale:} To maintain controlled experimental settings, our dataset is relatively small. Although sufficient for initial analysis, the limited size constrains the statistical power and the diversity of discourse contexts represented. Expanding the dataset will be essential to ensure robustness, enable richer cross-linguistic comparisons, and support generalization to real-world embodied scenarios.

\textbf{Cross-linguistic variation:} Our study focuses on English and Chinese, but demonstrative systems differ widely across languages. Some, such as Spanish, employ three contrasts rather than a simple proximal–distal distinction. Extending the dataset to more languages will allow us to capture richer cultural and typological variation.

\section*{Ethics Statement}

All data used in this study are constructed by the authors and do not involve copyright concerns. The dataset contains no offensive or harmful content. Ethical approval for the study is obtained from the Research Committee of the Department of Language Science and Technology of the Hong Kong Polytechnic University. AI assistants (e.g., ChatGPT, Copilot) are used only for language polishing and improving readability of the manuscript. They are not involved in the design of experiments, data collection, analysis, or interpretation of results. All scientific contributions and conclusions are solely those of the authors.

\section*{Informed Consent Statement}

Informed consent is obtained online from all participants prior to data collection. English‑speaking participants complete consent via Google Forms, and Chinese‑speaking participants via Credamo. All participants are over 18 years old and take part voluntarily during Nov 2025.

The consent procedure follows approved ethical guidelines. Participants are informed about the study’s purpose, procedures, data usage, and their right to withdraw at any time. No personally identifiable information is collected, and all responses are anonymized and handled confidentially. The study involves minimal risk and includes no vulnerable populations. It consists solely of non‑interventional, survey‑based data collection, and participants are assured that anonymized findings will be used only for academic research.

Full instructions provided to participants are as follows: \begin{quote}
    
Thank you very much for participating in this survey. We are conducting an academic study on commonsense. Please select the answers that feel most appropriate to you. Some questions may appear in slightly different forms. Please pay attention to the subtle differences. We will not collect any personally identifiable information. You may withdraw from the study at any time. The data collected in this survey will be used solely for academic research and publication purposes. This study involves minimal risk and consists only of non‑sensitive survey questions. To ensure response quality, we have included a few attention‑check items. Please avoid rushing through the survey. Thank you for your understanding and cooperation. The survey contains approximately 20 questions and is expected to take about 5 minutes to complete.\end{quote}

\section*{Potential Risks}
This study presents minimal risk. No sensitive personal information is collected, and all tasks involve non-‐ sensitive linguistic judgements.  Participants could choose to withdraw at any time without consequence.

\section*{Participant Recruitment and Compensation}
Chinese participants are recruited through Credamo, and English participants are recruited through Prolific. All participants receive monetary compensation at the standard rates recommended by each platform at the time of data collection. These rates are designed to be fair and appropriate for participants’ respective regions and demographics, and they align with the platforms’ minimum payment guidelines.

\section*{Acknowledgements}

EC was supported by a GRF grant from the Research Grants Council of the Hong Kong Special Administrative Region, China (Project No. PolyU 15601325).

\bibliography{custom}

\appendix

\section{Appendix}
\label{sec:appendix}
\subsection{Prompt Selection}
To establish a comprehensive evaluation framework, we begin with a preliminary comparison of five commonly used prompting strategies: (1) zero-shot, (2) zero-shot with prompt refinement, (3) zero-shot with chain-of-thought (CoT), (4) few-shot prompting, and (5) role-play prompting \citep{brown2020fewshot,kojima2022zeroshot,kong2023roleplay, xue2026reasonneededefficientgenerative}. Examples for each strategy are presented below:
\begin{itemize}
  \item \textbf{Zero-shot}: \textit{There's a bench. Ava and Irie are standing at the both sides. You are Ava. There are three books on the bench. You say, “Hide that  \{book\}.” Done! Are there any books left on the bench? If so, which ones? Option: 1. The one near Ava; 2. The one near Irie; 3. The one in the middle; 4. All of the above. }

  \item \textbf{Zero-shot prompt refinement}: Same as zero-shot, with: \textit{Please reply [The answer is: \{1, 2, 3, 4\}], and give a brief reason.}

  \item \textbf{Zero-shot CoT}: Same as zero-shot, with: \textit{Let’s think step by step.}

  \item \textbf{Few-shot}: An prior example with answers precede the target question.

  \item \textbf{Role-play}: Adds context: \textit{You are a household robot.} Then asks the same question.
\end{itemize}

From the original dataset, we select a balanced subset of 20 questions and evaluate five prompting strategies across five models in both English and Chinese. Each strategy–model–language combination is tested with three independent runs, yielding a total of 3000 responses in the preliminary analysis.

\begin{table*}[htbp]
\centering
\small
\resizebox{\textwidth}{!}{
\begin{tabular}{lllccccccc}
\toprule
\textbf{Model} & \textbf{Language} & \textbf{Strategy} & \textbf{1} & \textbf{1,2} & \textbf{1,3} & \textbf{2} & \textbf{2,3} & \textbf{3} & \textbf{4} \\
\midrule

\multirow{10}{*}{claude-sonnet-4.5}
& Chinese & few-shots & 1.67\% & 1.67\% & 38.33\% & 6.67\% & 26.67\% & 1.67\% & 23.33\% \\
& Chinese & role-play & 5.00\% & 0.00\% & 31.67\% & 5.00\% & 21.67\% & 1.67\% & 35.00\% \\
& Chinese & zero-shot & 5.00\% & 1.67\% & 23.33\% & 5.00\% & 21.67\% & 0.00\% & 43.33\% \\
& Chinese & zero-shot CoT & 3.33\% & 6.67\% & 33.33\% & 3.33\% & 23.33\% & 0.00\% & 30.00\% \\
& Chinese & zero-shot prompt refinement & 5.00\% & 1.67\% & 35.00\% & 3.33\% & 16.67\% & 0.00\% & 38.33\% \\
& English & few-shots & 1.67\% & 0.00\% & 31.67\% & 1.67\% & 16.67\% & 0.00\% & 48.33\% \\
& English & role-play & 1.67\% & 1.67\% & 28.33\% & 0.00\% & 25.00\% & 0.00\% & 43.33\% \\
& English & zero-shot & 3.33\% & 0.00\% & 20.00\% & 0.00\% & 18.33\% & 0.00\% & 58.33\% \\
& English & zero-shot CoT & 3.33\% & 0.00\% & 25.00\% & 1.67\% & 20.00\% & 3.33\% & 46.67\% \\
& English & zero-shot prompt refinement & 0.00\% & 0.00\% & 15.00\% & 0.00\% & 15.00\% & 1.67\% & 68.33\% \\

\midrule
\multirow{10}{*}{deepseek-chat-v3.1}
& Chinese & few-shots & 3.33\% & 1.67\% & 21.67\% & 1.67\% & 18.33\% & 6.67\% & 46.67\% \\
& Chinese & role-play & 3.33\% & 1.67\% & 11.67\% & 1.67\% & 18.33\% & 5.00\% & 58.33\% \\
& Chinese & zero-shot & 5.00\% & 1.67\% & 21.67\% & 0.00\% & 13.33\% & 0.00\% & 58.33\% \\
& Chinese & zero-shot CoT & 1.67\% & 1.67\% & 33.33\% & 0.00\% & 18.33\% & 0.00\% & 45.00\% \\
& Chinese & zero-shot prompt refinement & 1.67\% & 1.67\% & 25.00\% & 1.67\% & 23.33\% & 5.00\% & 41.67\% \\
& English & few-shots & 0.00\% & 0.00\% & 26.67\% & 0.00\% & 28.33\% & 0.00\% & 45.00\% \\
& English & role-play & 0.00\% & 0.00\% & 21.67\% & 1.67\% & 25.00\% & 0.00\% & 51.67\% \\
& English & zero-shot & 3.33\% & 0.00\% & 18.33\% & 0.00\% & 36.67\% & 0.00\% & 41.67\% \\
& English & zero-shot CoT & 1.67\% & 0.00\% & 28.33\% & 3.33\% & 26.67\% & 0.00\% & 40.00\% \\
& English & zero-shot prompt refinement & 1.67\% & 0.00\% & 30.00\% & 1.67\% & 30.00\% & 0.00\% & 36.67\% \\

\midrule
\multirow{10}{*}{gemini-2.5-pro}
& Chinese & few-shots & 3.33\% & 0.00\% & 26.67\% & 3.33\% & 41.67\% & 0.00\% & 25.00\% \\
& Chinese & role-play & 8.33\% & 0.00\% & 21.67\% & 0.00\% & 40.00\% & 1.67\% & 28.33\% \\
& Chinese & zero-shot & 5.00\% & 0.00\% & 33.33\% & 0.00\% & 40.00\% & 0.00\% & 21.67\% \\
& Chinese & zero-shot CoT & 3.33\% & 0.00\% & 33.33\% & 1.67\% & 38.33\% & 1.67\% & 21.67\% \\
& Chinese & zero-shot prompt refinement & 1.67\% & 0.00\% & 43.33\% & 0.00\% & 28.33\% & 3.33\% & 23.33\% \\
& English & few-shots & 6.67\% & 0.00\% & 30.00\% & 0.00\% & 38.33\% & 5.00\% & 20.00\% \\
& English & role-play & 5.00\% & 0.00\% & 36.67\% & 1.67\% & 41.67\% & 0.00\% & 15.00\% \\
& English & zero-shot & 0.00\% & 0.00\% & 36.67\% & 0.00\% & 43.33\% & 3.33\% & 16.67\% \\
& English & zero-shot CoT & 1.67\% & 0.00\% & 40.00\% & 1.67\% & 41.67\% & 1.67\% & 13.33\% \\
& English & zero-shot prompt refinement & 3.33\% & 0.00\% & 40.00\% & 1.67\% & 48.33\% & 0.00\% & 6.67\% \\

\midrule
\multirow{10}{*}{gpt-5.1}
& Chinese & few-shots & 1.67\% & 3.33\% & 23.33\% & 1.67\% & 23.33\% & 0.00\% & 46.67\% \\
& Chinese & role-play & 6.67\% & 5.00\% & 15.00\% & 3.33\% & 26.67\% & 0.00\% & 43.33\% \\
& Chinese & zero-shot & 3.33\% & 1.67\% & 18.33\% & 0.00\% & 28.33\% & 5.00\% & 43.33\% \\
& Chinese & zero-shot CoT & 1.67\% & 1.67\% & 31.67\% & 0.00\% & 21.67\% & 0.00\% & 43.33\% \\
& Chinese & zero-shot prompt refinement & 6.67\% & 1.67\% & 18.33\% & 0.00\% & 18.33\% & 1.67\% & 53.33\% \\
& English & few-shots & 1.67\% & 0.00\% & 23.33\% & 1.67\% & 26.67\% & 3.33\% & 43.33\% \\
& English & role-play & 1.67\% & 0.00\% & 15.00\% & 0.00\% & 15.00\% & 1.67\% & 66.67\% \\
& English & zero-shot & 5.00\% & 1.67\% & 15.00\% & 1.67\% & 26.67\% & 1.67\% & 48.33\% \\
& English & zero-shot CoT & 3.33\% & 0.00\% & 18.33\% & 1.67\% & 28.33\% & 0.00\% & 48.33\% \\
& English & zero-shot prompt refinement & 3.33\% & 0.00\% & 11.67\% & 1.67\% & 15.00\% & 0.00\% & 68.33\% \\

\midrule
\multirow{10}{*}{qwen3-max}
& Chinese & few-shots & 25.00\% & 3.33\% & 13.33\% & 8.33\% & 15.00\% & 0.00\% & 35.00\% \\
& Chinese & role-play & 33.33\% & 0.00\% & 8.33\% & 11.67\% & 11.67\% & 0.00\% & 35.00\% \\
& Chinese & zero-shot & 26.67\% & 5.00\% & 18.33\% & 6.67\% & 20.00\% & 0.00\% & 23.33\% \\
& Chinese & zero-shot CoT & 36.67\% & 5.00\% & 1.67\% & 8.33\% & 13.33\% & 0.00\% & 35.00\% \\
& Chinese & zero-shot prompt refinement & 38.33\% & 3.33\% & 11.67\% & 1.67\% & 16.67\% & 1.67\% & 26.67\% \\
& English & few-shots & 21.67\% & 0.00\% & 13.33\% & 8.33\% & 20.00\% & 1.67\% & 35.00\% \\
& English & role-play & 16.67\% & 0.00\% & 13.33\% & 6.67\% & 20.00\% & 0.00\% & 43.33\% \\
& English & zero-shot & 15.00\% & 3.33\% & 16.67\% & 3.33\% & 15.00\% & 0.00\% & 46.67\% \\
& English & zero-shot CoT & 20.00\% & 1.67\% & 18.33\% & 3.33\% & 23.33\% & 1.67\% & 31.67\% \\
& English & zero-shot prompt refinement & 16.67\% & 0.00\% & 13.33\% & 10.00\% & 13.33\% & 1.67\% & 45.00\% \\

\bottomrule
\end{tabular}
}
\caption{Performance across models, languages, and prompting strategies.}
\label{tab:prompt}
\end{table*}

The results are shown in Table \ref{tab:prompt}. Across models and languages, the observed differences between prompting strategies are small and inconsistent. No strategy demonstrates a systematic advantage on the key performance dimensions relevant to our study. In particular, strategies such as few-shot, role-play, and zero-shot variants produce fluctuations that are within the range of run-to-run variability rather than reflecting stable performance differences. These findings indicate that prompting style is not a major determinant of model outcomes for this task.

Given the minimal impact of prompting strategy, we adopt a zero-shot evaluation framework for the main study. This choice aligns with how humans naturally approach the task—responding directly without exposure to examples—and therefore provides a more ecologically valid comparison between human and model performance. Among the zero-shot variants, we select the refined version to standardize response format and reduce ambiguity in interpretation. The refined prompt also requires participants to provide structured answers with brief explanations, which improves the clarity and interpretability of the collected responses.

\subsection{Results of the Control Group}
The results are presented in Table \ref{tab:control_group_1} to Table \ref{tab:control_group_6}.

\begin{table*}[ht]
\centering
\small
\begin{tabular}{llcccccccc}
\hline
Model & Class & 1 & 1,2 & 1,3 & 2 & 2,3 & 3 & 4 & None \\
\hline
claude-sonnet-4.5 & Self: Proximal   & / & 6.00\% & 2.00\% & 6.00\% & 56.00\% & / & 30.00\% & / \\
                  & Self: Distal   & 16.00\% & 28.00\% & 54.00\% & / & / & / & 2.00\% & / \\
                  & Other: Proximal  & 16.00\% & 16.00\% & 54.00\% & / & / & / & 14.00\% & / \\
                  & Other: Distal  & / & / & / & 6.00\% & 94.00\% & / & / & / \\
\hline
deepseek-v3.1     & Self: Proximal   & / & 40.00\% & / & 20.00\% & 14.00\% & 4.00\% & 18.00\% & 4.00\% \\
                  & Self: Distal   & 10.00\% & 48.00\% & 8.00\% & 6.00\% & / & / & 28.00\% & / \\
                  & Other: Proximal  & 10.00\% & 60.00\% & 12.00\% & / & / & / & 16.00\% & 2.00\% \\
                  & Other: Distal  & / & 34.00\% & / & 6.00\% & 10.00\% & / & 50.00\% & / \\
\hline
gemini-2.5-pro    & Self: Proximal   & / & / & / & 2.00\% & 48.00\% & 4.00\% & 46.00\% & / \\
                  & Self: Distal   & 8.00\% & / & 26.00\% & / & / & 14.00\% & 50.00\% & 2.00\% \\
                  & Other: Proximal  & 6.00\% & / & 32.00\% & / & / & 2.00\% & 56.00\% & 4.00\% \\
                  & Other: Distal  & / & / & / & 10.00\% & 46.00\% & 2.00\% & 42.00\% & / \\
\hline
gpt5.1            & Self: Proximal   & / & / & / & 10.00\% & 58.00\% & / & 30.00\% & 2.00\% \\
                  & Self: Distal   & 14.00\% & / & 30.00\% & 2.00\% & / & 42.00\% & 12.00\% & / \\
                  & Other: Proximal  & 22.00\% & / & 50.00\% & / & / & / & 28.00\% & / \\
                  & Other: Distal  & / & / & / & 18.00\% & 58.00\% & 10.00\% & 12.00\% & 2.00\% \\
\hline
qwen3-max         & Self: Proximal   & / & / & / & 8.00\% & 78.00\% & 2.00\% & 12.00\% & / \\
                  & Self: Distal   & 14.00\% & 6.00\% & 22.00\% & 8.00\% & / & 38.00\% & 12.00\% & / \\
                  & Other: Proximal  & 10.00\% & 2.00\% & 58.00\% & 2.00\% & 4.00\% & 12.00\% & 12.00\% & / \\
                  & Other: Distal  & / & / & / & / & 58.00\% & 40.00\% & 2.00\% & / \\
\hline
human             & Self: Proximal   & / & / & 1.00\% & 0.50\% & 95.50\% & 1.00\% & 2.00\% & / \\
                  & Self: Distal   & / & / & 92.00\% & / & 6.00\% & 1.00\% & 1.00\% & / \\
                  & Other: Proximal  & / & / & 81.00\% & / & 18.50\% & / & 0.50\% & / \\
                  & Other: Distal  & / & / & 19.00\% & / & 80.00\% & / & 1.00\% & / \\
\hline
\end{tabular}
\caption{Comparative results of models and human performance in English under the only-pronoun condition}
\label{tab:control_group_1}
\end{table*}

\begin{table*}[htbp]
\centering
\small
\begin{tabular}{llcccccccc}
\hline
Model & Class & 1 & 1,2 & 1,3 & 2 & 2,3 & 3 & 4 & None \\
\hline
claude-sonnet-4.5 & Self: Proximal   & / & / & / & 100.00\% & / & / & / & / \\
                  & Self: Distal   & 60.00\% & / & 4.00\% & 16.00\% & / & 20.00\% & / & / \\
                  & Other: Proximal  & 62.00\% & / & 4.00\% & / & / & 14.00\% & 20.00\% & / \\
                  & Other: Distal  & / & / & / & 100.00\% & / & / & / & / \\
\hline
deepseek-v3.1     & Self: Proximal   & 8.00\% & / & / & 22.00\% & / & 16.00\% & 50.00\% & 4.00\% \\
                  & Self: Distal   & 18.00\% & 4.00\% & 2.00\% & 20.00\% & / & 6.00\% & 50.00\% & / \\
                  & Other: Proximal  & 2.00\% & 4.00\% & 2.00\% & 24.00\% & / & 6.00\% & 62.00\% & / \\
                  & Other: Distal  & 16.00\% & / & / & 22.00\% & 6.00\% & 8.00\% & 48.00\% & / \\
\hline
gemini-2.5-pro    & Self: Proximal   & 2.00\% & / & / & 12.00\% & 62.00\% & 4.00\% & 20.00\% & / \\
                  & Self: Distal   & 44.00\% & / & 34.00\% & / & / & 4.00\% & 12.00\% & 6.00\% \\
                  & Other: Proximal  & 20.00\% & / & 56.00\% & 2.00\% & / & 2.00\% & 18.00\% & 2.00\% \\
                  & Other: Distal  & / & / & / & 14.00\% & 54.00\% & / & 30.00\% & 2.00\% \\
\hline
gpt5.1            & Self: Proximal   & / & / & / & 12.00\% & 66.00\% & / & 22.00\% & / \\
                  & Self: Distal   & 14.00\% & / & 68.00\% & / & 2.00\% & 4.00\% & 12.00\% & / \\
                  & Other: Proximal  & 10.00\% & 2.00\% & 64.00\% & / & / & / & 22.00\% & 2.00\% \\
                  & Other: Distal  & / & / & / & 22.00\% & 60.00\% & / & 18.00\% & / \\
\hline
qwen3-max         & Self: Proximal   & / & / & / & 16.00\% & 80.00\% & / & 4.00\% & / \\
                  & Self: Distal   & 14.00\% & / & 10.00\% & 14.00\% & 6.00\% & 48.00\% & 8.00\% & / \\
                  & Other: Proximal  & 16.00\% & 2.00\% & 56.00\% & 8.00\% & 2.00\% & 8.00\% & 8.00\% & / \\
                  & Other: Distal  & / & / & / & 20.00\% & 72.00\% & 2.00\% & 6.00\% & / \\
\hline
human             & Self: Proximal   & 0.50\% & / & 0.00\% & / & 100.00\% & / & / & / \\
                  & Self: Distal   & 0.50\% & / & 98.00\% & / & 2.00\% & / & / & / \\
                  & Other: Proximal  & / & 0.50\% & 97.50\% & / & 2.00\% & / & / & / \\
                  & Other: Distal  & / & 0.50\% & 1.50\% & / & 98.00\% & / & / & / \\
\hline
\end{tabular}
\caption{Comparison across models and humans in Chinese for the only-pronoun condition}
\end{table*}

\begin{table*}[htbp]
\centering
\small
\begin{tabular}{llcccccccc}
\hline
Model & Class & 1 & 1,2 & 1,3 & 2 & 2,3 & 3 & 4 & None \\
\hline
claude-sonnet-4.5 & Self: Proximal   & / & 4.00\% & / & 2.00\% & 60.00\% & / & 32.00\% & 2.00\% \\
                  & Self: Distal   & 14.00\% & / & 80.00\% & / & / & / & 6.00\% & / \\
                  & Other: Proximal  & 4.00\% & 6.00\% & 60.00\% & / & / & / & 28.00\% & 2.00\% \\
                  & Other: Distal  & / & / & / & 2.00\% & 92.00\% & / & 4.00\% & 2.00\% \\
\hline
deepseek-v3.1     & Self: Proximal   & / & 40.00\% & / & 6.00\% & 18.00\% & / & 34.00\% & 2.00\% \\
                  & Self: Distal   & 4.00\% & 44.00\% & 18.00\% & 8.00\% & / & / & 26.00\% & / \\
                  & Other: Proximal  & 2.00\% & 52.00\% & 10.00\% & 2.00\% & / & / & 30.00\% & 4.00\% \\
                  & Other: Distal  & / & 48.00\% & / & 8.00\% & 16.00\% & / & 28.00\% & / \\
\hline
gemini-2.5-pro    & Self: Proximal   & / & / & / & 4.00\% & 36.00\% & 6.00\% & 54.00\% & / \\
                  & Self: Distal   & 14.00\% & / & 48.00\% & / & / & / & 38.00\% & / \\
                  & Other: Proximal  & 6.00\% & / & 50.00\% & / & / & 2.00\% & 40.00\% & 2.00\% \\
                  & Other: Distal  & 2.00\% & / & / & 4.00\% & 60.00\% & 2.00\% & 30.00\% & 2.00\% \\
\hline
gpt5.1            & Self: Proximal   & / & / & / & 2.00\% & 62.00\% & / & 36.00\% & / \\
                  & Self: Distal   & 28.00\% & / & 56.00\% & / & / & 4.00\% & 10.00\% & 2.00\% \\
                  & Other: Proximal  & 22.00\% & / & 26.00\% & / & / & / & 46.00\% & 6.00\% \\
                  & Other: Distal  & / & / & / & 12.00\% & 60.00\% & / & 28.00\% & / \\
\hline
qwen3-max         & Self: Proximal   & / & / & / & / & 22.00\% & / & 78.00\% & / \\
                  & Self: Distal   & 8.00\% & 4.00\% & 52.00\% & 6.00\% & 6.00\% & 18.00\% & 6.00\% & / \\
                  & Other: Proximal  & 2.00\% & 2.00\% & 22.00\% & / & / & 50.00\% & 24.00\% & / \\
                  & Other: Distal  & / & / & / & 4.00\% & 74.00\% & 8.00\% & 14.00\% & / \\
\hline
human             & Self: Proximal   & / & / & 1.00\% & 1.00\% & 97.00\% & 0.50\% & 0.50\% & / \\
                  & Self: Distal   & 1.00\% & / & 89.50\% & 1.50\% & 7.00\% & 0.50\% & 0.50\% & / \\
                  & Other: Proximal  & / & / & 69.50\% & 0.50\% & 28.50\% & 0.50\% & 1.00\% & / \\
                  & Other: Distal  & 1.00\% & / & 30.00\% & / & 67.50\% & 0.50\% & 1.00\% & / \\
\hline
\end{tabular}
\caption{Comparison across models and humans in English for the reinforcing-demonstratives-pronouns condition}
\end{table*}

\begin{table*}[htbp]
\centering
\small
\begin{tabular}{llcccccccc}
\hline
Model & Class & 1 & 1,2 & 1,3 & 2 & 2,3 & 3 & 4 & None \\
\hline
claude-sonnet-4.5 & Self: Proximal   & / & / & / & 88.00\% & / & / & 12.00\% & / \\
                  & Self: Distal   & 22.00\% & / & 40.00\% & / & / & 38.00\% & / & / \\
                  & Other: Proximal  & 30.00\% & / & 36.00\% & 4.00\% & / & 30.00\% & / & / \\
                  & Other: Distal  & / & / & / & 100.00\% & / & / & / & / \\
\hline
deepseek-v3.1     & Self: Proximal   & 12.00\% & 2.00\% & / & 16.00\% & 6.00\% & 20.00\% & 42.00\% & 2.00\% \\
                  & Self: Distal   & 12.00\% & 2.00\% & 6.00\% & 30.00\% & / & 10.00\% & 38.00\% & / \\
                  & Other: Proximal  & 8.00\% & 4.00\% & 4.00\% & 14.00\% & 2.00\% & 6.00\% & 62.00\% & / \\
                  & Other: Distal  & 18.00\% & / & / & 18.00\% & 4.00\% & 2.00\% & 58.00\% & / \\
\hline
gemini-2.5-pro    & Self: Proximal   & / & / & / & 16.00\% & 64.00\% & 2.00\% & 16.00\% & 2.00\% \\
                  & Self: Distal   & 20.00\% & / & 48.00\% & / & / & 2.00\% & 22.00\% & 8.00\% \\
                  & Other: Proximal  & 18.00\% & / & 56.00\% & 2.00\% & / & / & 14.00\% & 10.00\% \\
                  & Other: Distal  & / & / & / & 12.00\% & 68.00\% & / & 18.00\% & 2.00\% \\
\hline
gpt5.1            & Self: Proximal   & 2.00\% & / & / & 2.00\% & 74.00\% & / & 22.00\% & / \\
                  & Self: Distal   & 6.00\% & / & 86.00\% & 2.00\% & / & / & 6.00\% & / \\
                  & Other: Proximal  & 8.00\% & / & 80.00\% & / & / & / & 12.00\% & / \\
                  & Other: Distal  & / & / & / & 2.00\% & 94.00\% & / & 4.00\% & / \\
\hline
qwen3-max         & Self: Proximal   & / & / & / & / & 100.00\% & / & / & / \\
                  & Self: Distal   & 2.00\% & / & 38.00\% & 14.00\% & 10.00\% & 36.00\% & / & / \\
                  & Other: Proximal  & / & / & 78.00\% & 14.00\% & / & 8.00\% & / & / \\
                  & Other: Distal  & / & / & / & 2.00\% & 98.00\% & / & / & / \\
\hline
human             & Self: Proximal   & / & / & / & / & 100.00\% & / & / & / \\
                  & Self: Distal   & / & / & 98.00\% & / & 2.00\% & / & / & / \\
                  & Other: Proximal  & / & / & 100.00\% & / & / & / & / & / \\
                  & Other: Distal  & / & / & / & / & 99.50\% & 0.50\% & / & / \\
\hline
\end{tabular}
\caption{Comparison across models and humans in Chinese for the reinforcing-demonstratives-pronouns condition}
\end{table*}

\begin{table*}[htbp]
\centering
\small
\begin{tabular}{llcccccccc}
\hline
Model & Class & 1 & 1,2 & 1,3 & 2 & 2,3 & 3 & 4 & None \\
\hline
claude-sonnet-4.5 & Self: Proximal   & 10.00\% & 2.00\% & 60.00\% & / & / & / & 26.00\% & 2.00\% \\
                  & Self: Distal   & 2.00\% & 4.00\% & 2.00\% & 8.00\% & 34.00\% & / & 50.00\% & / \\
                  & Other: Proximal  & / & 4.00\% & 10.00\% & 2.00\% & 72.00\% & / & 12.00\% & / \\
                  & Other: Distal  & 10.00\% & 8.00\% & 24.00\% & 2.00\% & 2.00\% & 2.00\% & 48.00\% & 4.00\% \\
\hline
deepseek-v3.1     & Self: Proximal   & 8.00\% & 48.00\% & 16.00\% & 8.00\% & / & / & 20.00\% & / \\
                  & Self: Distal   & / & 36.00\% & / & 6.00\% & 24.00\% & / & 30.00\% & 4.00\% \\
                  & Other: Proximal  & 2.00\% & 46.00\% & / & 4.00\% & 20.00\% & / & 28.00\% & / \\
                  & Other: Distal  & 6.00\% & 48.00\% & 14.00\% & 6.00\% & 2.00\% & 2.00\% & 20.00\% & 2.00\% \\
\hline
gemini-2.5-pro    & Self: Proximal   & / & 2.00\% & 18.00\% & / & 16.00\% & 12.00\% & 50.00\% & 2.00\% \\
                  & Self: Distal   & 44.00\% & / & 6.00\% & 4.00\% & 4.00\% & / & 42.00\% & / \\
                  & Other: Proximal  & / & / & 4.00\% & 8.00\% & 36.00\% & 2.00\% & 50.00\% & / \\
                  & Other: Distal  & 4.00\% & / & 24.00\% & / & 8.00\% & 2.00\% & 62.00\% & / \\
\hline
gpt5.1            & Self: Proximal   & 10.00\% & / & 18.00\% & 10.00\% & 14.00\% & 4.00\% & 38.00\% & 6.00\% \\
                  & Self: Distal   & 18.00\% & 2.00\% & 30.00\% & 4.00\% & 2.00\% & / & 40.00\% & 4.00\% \\
                  & Other: Proximal  & 6.00\% & / & 24.00\% & 4.00\% & 8.00\% & 4.00\% & 52.00\% & 2.00\% \\
                  & Other: Distal  & 4.00\% & / & 18.00\% & 8.00\% & 26.00\% & / & 36.00\% & 8.00\% \\
\hline
qwen3-max         & Self: Proximal   & 4.00\% & 2.00\% & 24.00\% & 14.00\% & 16.00\% & 18.00\% & 22.00\% & / \\
                  & Self: Distal   & / & / & / & / & 36.00\% & 2.00\% & 62.00\% & / \\
                  & Other: Proximal  & / & / & / & 6.00\% & 48.00\% & 22.00\% & 24.00\% & / \\
                  & Other: Distal  & 2.00\% & 2.00\% & 26.00\% & / & 2.00\% & 40.00\% & 28.00\% & / \\
\hline
human             & Self: Proximal   & 3.50\% & / & 77.50\% & 1.00\% & 16.00\% & 1.50\% & 0.50\% & / \\
                  & Self: Distal   & / & 1.00\% & 3.50\% & 2.50\% & 92.50\% & / & 0.50\% & / \\
                  & Other: Proximal  & 1.50\% & / & 28.50\% & 1.50\% & 68.50\% & / & / & / \\
                  & Other: Distal  & 2.50\% & 2.00\% & 67.50\% & / & 28.00\% & / & / & / \\
\hline
\end{tabular}
\caption{Comparison across models and humans in English for the inconsistent-demonstratives-pronouns condition}
\end{table*}
\begin{table*}[htbp]
\centering
\small
\begin{tabular}{llcccccccc}
\hline
Model & Class & 1 & 1,2 & 1,3 & 2 & 2,3 & 3 & 4 & None \\
\hline
claude-sonnet-4.5 & Self: Proximal   & 50.00\% & / & 24.00\% & / & / & 26.00\% & / & / \\
                  & Self: Distal   & / & / & / & 78.00\% & 2.00\% & / & 20.00\% & / \\
                  & Other: Proximal  & / & / & / & 100.00\% & / & / & / & / \\
                  & Other: Distal  & 44.00\% & / & 36.00\% & / & / & 20.00\% & / & / \\
\hline
deepseek-v3.1     & Self: Proximal   & 12.00\% & 2.00\% & 4.00\% & 30.00\% & / & 8.00\% & 44.00\% & / \\
                  & Self: Distal   & 8.00\% & 4.00\% & 2.00\% & 34.00\% & 2.00\% & 4.00\% & 46.00\% & / \\
                  & Other: Proximal  & 4.00\% & / & / & 30.00\% & 2.00\% & 4.00\% & 60.00\% & / \\
                  & Other: Distal  & 2.00\% & 12.00\% & 2.00\% & 20.00\% & / & 12.00\% & 52.00\% & / \\
\hline
gemini-2.5-pro    & Self: Proximal   & 10.00\% & / & 54.00\% & 4.00\% & / & 6.00\% & 26.00\% & / \\
                  & Self: Distal   & / & / & / & 18.00\% & 54.00\% & 4.00\% & 24.00\% & / \\
                  & Other: Proximal  & / & / & / & 12.00\% & 56.00\% & / & 30.00\% & 2.00\% \\
                  & Other: Distal  & 18.00\% & / & 34.00\% & 4.00\% & / & 2.00\% & 34.00\% & 8.00\% \\
\hline
gpt5.1            & Self: Proximal   & 2.00\% & / & 90.00\% & / & / & / & 6.00\% & 2.00\% \\
                  & Self: Distal   & / & / & / & 4.00\% & 66.00\% & / & 30.00\% & / \\
                  & Other: Proximal  & / & / & 2.00\% & 4.00\% & 82.00\% & 2.00\% & 10.00\% & / \\
                  & Other: Distal  & 8.00\% & / & 80.00\% & / & / & / & 10.00\% & 2.00\% \\
\hline
qwen3-max         & Self: Proximal   & / & / & 4.00\% & 22.00\% & 40.00\% & 28.00\% & 6.00\% & / \\
                  & Self: Distal   & / & 2.00\% & / & 12.00\% & 86.00\% & / & / & / \\
                  & Other: Proximal  & / & / & / & 4.00\% & 90.00\% & 4.00\% & 2.00\% & / \\
                  & Other: Distal  & 4.00\% & / & 84.00\% & 4.00\% & 2.00\% & 6.00\% & / & / \\
\hline
human             & Self: Proximal   & 0.50\% & / & 93.00\% & / & 6.50\% & / & / & / \\
                  & Self: Distal   & / & / & / & 1.00\% & 99.00\% & / & / & / \\
                  & Other: Proximal  & / & / & 2.00\% & 0.50\% & 97.50\% & / & / & / \\
                  & Other: Distal  & 1.00\% & / & 98.50\% & / & 0.50\% & / & / & / \\
\hline
\end{tabular}
\caption{Comparison across models and humans in Chinese for the inconsistent-demonstratives-pronouns condition}
\label{tab:control_group_6}
\end{table*}

\end{CJK}
\end{document}